\pdfoutput=1

\documentclass[11pt]{article}

\usepackage[final]{acl}

\usepackage{times}
\usepackage{latexsym}

\usepackage[T1]{fontenc}

\usepackage[utf8]{inputenc}

\usepackage{microtype}

\usepackage{inconsolata}

\usepackage{hyperref}       
\usepackage{url}            
\usepackage{booktabs}       
\usepackage{amsfonts}       
\usepackage{nicefrac}       
\usepackage{microtype}      
\usepackage{xcolor}         
\usepackage{xspace}

\usepackage{natbib}
\usepackage{enumitem} 
\usepackage{xspace}
\usepackage{graphicx}
\usepackage{tabularx}
\usepackage{colortbl}
\usepackage{xcolor}
\usepackage{array}
\usepackage{multirow}
\usepackage{soul}
\usepackage{hyperref}
\usepackage{balance}
\usepackage{graphicx}
\usepackage{subcaption}
\usepackage{graphicx}
\usepackage{xspace}
\usepackage{makecell}

\newcommand{\method}{{NLP-ADBench}\xspace}

\title{\method: NLP Anomaly Detection Benchmark}

\author{
  \textbf{Yuangang Li\textsuperscript{1,*}}, 
  \textbf{Jiaqi Li\textsuperscript{1,*}}, 
  \textbf{Zhuo Xiao\textsuperscript{1,*}}, 
  \textbf{Tiankai Yang\textsuperscript{1}}, 
  \textbf{Yi Nian\textsuperscript{1}}, 
  \textbf{Xiyang Hu\textsuperscript{2\textsuperscript{†}}}, 
  \textbf{Yue Zhao\textsuperscript{1\textsuperscript{†}}} \\
  \textsuperscript{1}University of Southern California \quad
  \textsuperscript{2}Arizona State University \\
  \texttt{\{yuangang, jli77629, zhuoxiao, tiankaiy, yinian, yzhao010\}@usc.edu}, \\
  \texttt{xiyanghu@asu.edu} \\
  \textsuperscript{†}Corresponding authors
}

\begin{document}
\maketitle

\begin{abstract}
Anomaly detection (AD) is an important machine learning task with applications in fraud detection, content moderation, and user behavior analysis. However, AD is relatively understudied in a natural language processing (NLP) context, limiting its effectiveness in detecting harmful content, phishing attempts, and spam reviews.
We introduce \method, the most comprehensive NLP anomaly detection (NLP-AD) benchmark to date, which includes eight curated datasets and 19 state-of-the-art algorithms. These span 3 end-to-end methods and 16 two-step approaches that adapt classical, non-AD methods to language embeddings from BERT and OpenAI. 
Our empirical results show that no single model dominates across all datasets, indicating a need for automated model selection. Moreover, two-step methods with transformer-based embeddings consistently outperform specialized end-to-end approaches, with OpenAI embeddings outperforming those of BERT. 
We release \method at 
\url{https://github.com/USC-FORTIS/NLP-ADBench}
, providing a unified framework for NLP-AD and supporting future investigations. 
\end{abstract}

\section{Introduction}
Anomaly detection (AD) is a fundamental area in machine learning with diverse applications in web systems, such as fraud detection, content moderation, and user behavior analysis~\cite{chandola2009anomaly, ahmed2016survey}. 
Substantial progress has been achieved in AD for structured data such as tabular, graph, and time series~\cite{chalapathy2019deep,han2022adbench,lai2021revisiting,liu2022bond}, but its extension to natural language processing (NLP) remains relatively underexplored~\cite{ruff2021unifying,yang2024ad}. 
This gap limits our ability to identify harmful content, phishing attempts, and spam reviews.

For instance, detecting abusive or threatening language is crucial for ensuring that social media platforms and online forums remain safe environments for users~\cite{fortuna2018survey}. 
Likewise, detecting anomalous product reviews or descriptions in e-commerce is important for preserving user trust and platform credibility~\cite{chino2017voltime}. 
However, many standard AD methods are designed for \textit{numeric} or \textit{categorical} data and are not easily adapted to 
unstructured text~\cite{zhao_pyod_2019,chen_pyod_2024}. 
Existing studies on NLP-specific AD are limited in both dataset variety and algorithmic range \cite{han2022adbench,liu2022bond,yang2024ad}, leaving open questions about which approaches work best under different conditions.
These gaps lead to a central research question: 
\textit{How can we systematically evaluate and compare diverse AD methods
across real-world text datasets, and what insights can be gained to guide future development in NLP-based AD?}

\noindent
\textbf{Our Proposal and Key Contributions}. 
We introduce \method, the most comprehensive benchmark for NLP-AD tasks. 
\method offers four major benefits compared to prior work~\cite{bejan2023ad}:
(\textit{\textbf{i}}) eight real-world datasets covering a wide range of web use cases;
(\textit{\textbf{ii}}) 19 advanced methods that apply standard AD algorithms to language embeddings or use end-to-end neural architectures;
(\textit{\textbf{iii}}) detailed empirical findings that highlight new directions for NLP-AD;
and (\textit{\textbf{iv}}) fully open-source resources, including datasets, algorithm implementations, and more, aligns with the \textit{Resources and Evaluation} track.


\noindent
\textbf{Key Insights/Takeaways (see details in \S \ref{sec:experiments}).}  
Our comprehensive experiments reveal:
(\textit{\textbf{i}}) No single model dominates across all datasets, showing the need for model selection;
(\textit{\textbf{ii}}) Transformer-based embeddings substantially boost two-step AD methods (e.g., LUNAR \cite{goodge2022lunar} and LOF \cite{breunig2000lof}) relative to end-to-end approaches;
(\textit{\textbf{iii}}) High-dimensional embeddings (e.g., from OpenAI) improve detection performance, but also raise computational overhead;
and (\textit{\textbf{iv}}) Dataset-specific biases and human-centered anomaly definitions remain challenging for building robust and widely applicable NLP-AD systems.

\noindent

\noindent

\section{\method: AD Benchmark for NLP Tasks}
\vspace{-0.1in}
\subsection{Preliminaries and Problem Definition} \label{subsec:Preliminaries}
Anomaly Detection in Natural Language Processing (\textbf{NLP-AD}) focuses on identifying text instances that deviate significantly from expected or typical patterns. Unlike structured data, text data is inherently unstructured, high-dimensional, and deeply influenced by the nuances of human language, including syntax, semantics, and context~\cite{Aggarwal2017,yang2024ad}.
These unique properties introduce significant challenges, making the development of robust and accurate AD methods for NLP a complex and demanding task.

Formally, let $\mathcal{D} = {x_1, x_2, \dots, x_N}$ denote a corpus where each $x_i$ is a text instance. The goal of NLP-AD is to learn an anomaly scoring function $f:\mathcal{X} \rightarrow \mathbb{R}$ that assigns a real-valued anomaly score to each text instance. Higher scores denote greater deviations from normal patterns, indicating a higher likelihood of an anomalous instance.
\vspace{-0.1in}

\subsection{Curated Benchmark Datasets}
\label{subsec:datasets}

The limited availability of purpose-built datasets constrains the development and evaluation of effective methods in NLP-AD. To address this gap, we \textbf{curated} and \textbf{transformed} 8 existing classification datasets from various NLP domains into specialized datasets tailored for NLP-AD tasks, ensuring that all data are presented in a standard format. These datasets, collectively called the NLPAD datasets, provide a foundational resource for advancing research.

Each transformed dataset is named by adding the prefix ``NLPAD-" to the original dataset's name (e.g., NLPAD-AGNews, NLPAD-BBCNews), distinguishing them from the original datasets. The NLPAD datasets are provided in a unified JSON Lines format
for compatibility and ease of use. Each line is a JSON object
with four fields: text (the text used for anomaly detection), label (the anomaly detection label, where 1 represents an anomaly and 0 represents normal), original\_task (the task of the original dataset), and original\_label (the category label from the original dataset).

To transform each dataset for NLP-AD, we established a text selection process based on the data format. For tabular data, we carefully chose appropriate columns as the text source. 
For document-based data, we extracted text directly from relevant documents. The anomalous class for each dataset was selected based on \textit{semantic distinctions} within the dataset categories, ensuring that the identified anomalies represent meaningful deviations from the normal data distribution \cite{emmott2015meta,han2022adbench}. Once identified, the anomalous class was downsampled to represent less than 10\% of the total instances.

For instance, in the NLPAD-AGNews dataset (tabular data), we selected the ``description'' column as the text source, with the ``World'' category serving as the anomalous class due to its semantic divergence from other categories such as ``Sports'' or ``Technology.'' 
This anomalous class was then downsampled accordingly. Similarly, in the NLPAD-BBCNews dataset (document-based data), text from BBC News documents was used, with the ``entertainment'' category identified as anomalous because its semantic content significantly differs from other categories like ``Politics'' or ``Business.'' The ``entertainment'' category was also downsampled to maintain consistency. This semantic-driven approach to defining anomalies was consistently applied across all datasets. 
Further details of dataset sources and construction processes can be found in Appx. \ref{appx:datasets_sources}. Table.~\ref{tab:NLPAD_datasets_info_only_stat} presents the statistical information of the NLPAD datasets, including the total number of samples, the number of normal and anomalous samples, and the anomaly ratio for each dataset.

\vspace{-0.1in}
\begin{table}[!t]
\centering
\small 
\setlength{\tabcolsep}{2pt}
\vspace{-0.1in}
\caption{Statistical information of the NLPAD dataset.}
\vspace{-0.1in}
\label{tab:NLPAD_datasets_info_only_stat}

\scalebox{0.85}{
\begin{tabular}{l|cccc}
\hline
\textbf{NLPAD Dataset} & \textbf{\# Samples} & \textbf{\#Normal} & \textbf{\#Anomaly} & \textbf{\%Anomaly} \\
\hline
NLPAD-AGNews          & $98,207$ & $94,427$ & $3,780$ & $3.85\%$ \\
NLPAD-BBCNews           & $1,785$ & $1,723$ & $62$ & $3.47\%$ \\
NLPAD-EmailSpam          & $3,578$ & $3,432$ & $146$ & $4.08\%$ \\
NLPAD-Emotion           & $361,980$ & $350,166$ & $11,814$ & $3.26\%$ \\
NLPAD-MovieReview          & $26,369$ & $24,882$ & $1,487$ & $5.64\%$ \\
NLPAD-N24News         & $59,822$ & $57,994$ & $1,828$ & $3.06\%$ \\
NLPAD-SMSSpam          & $4,672$ & $4,518$ & $154$ & $3.30\%$ \\
NLPAD-YelpReview       & $316,924$ & $298,986$ & $17,938$ & $5.66\%$ \\
\hline
\end{tabular}}
\vspace{-0.2in}
\end{table}

\subsection{The Most Comprehensive NLP-AD Algorithms with Open Implementations}
\label{subsec:algorithms}

\noindent
Compared to the existing NLP-AD benchmark by Bejan et al.~\cite{bejan2023ad}, \method provides a broader evaluation by including 19 algorithms, categorized into two groups. 
The first group comprises 3 end-to-end algorithms that directly process raw text data to produce anomaly detection outcomes. The second group consists of 16 algorithms derived by applying 8 traditional anomaly detection (AD) methods to text embeddings generated from two models: \textit{bert-base-uncased}~\cite{devlin2019bert} and OpenAI’s \textit{text-embedding-3-large}~\cite{openai2024embedding}. These traditional AD methods do not operate on raw text directly but instead perform anomaly detection on embeddings, offering a complementary approach to the end-to-end methods. This comprehensive algorithm collection enables a robust evaluation of direct and embedding-based NLP anomaly detection techniques.
Here, we provide a brief description; see details in Appx. \ref{appx:algorithms}.

\noindent
\textbf{\textit{End-to-end NLP-AD Algorithms}.}
We evaluate 3 end-to-end algorithms tailored for NLP-AD. (1) \textbf{Context Vector Data Description (CVDD)}~\cite{ruff2019} leverages context vectors and pre-trained embeddings with a multi-head self-attention mechanism to project normal instances close to learned contexts, identifying anomalies based on deviations. (2) \textbf{Detecting Anomalies in Text via Self-Supervision of Transformers (DATE)}~\cite{manolache2021date} trains transformers using self-supervised tasks like replaced mask detection to capture normal text patterns and flag anomalies. (3) \textbf{Few-shot Anomaly Detection in Text with Deviation Learning (FATE)}~\cite{das2023fate} uses a few labeled anomalies with deviation learning to distinguish anomalies from normal instances. We adapt it to train solely on normal data, referring to the adapted version as FATE*.

\noindent
\textbf{\textit{Two-step NLP-AD Algorithms}.}
We evaluate 8 two-step algorithms that rely on embeddings generated by models such as \textit{bert-base-uncased}~\cite{devlin2019bert} and \textit{text-embedding-3-large}~\cite{openai2024embedding}.
These algorithms are designed to work with structured numerical data and cannot directly process raw textual data, requiring text transformation into numerical embeddings. 
(4) \textbf{LOF}~\cite{breunig2000lof} measures local density deviations, while (5) \textbf{DeepSVDD}~\cite{ruff2018deep} minimizes the volume of a hypersphere enclosing normal representations. 
(6) \textbf{ECOD}~\cite{li2022ecod} uses empirical cumulative distribution functions to estimate densities and assumes anomalies lie in distribution tails. 
(7) \textbf{IForest}~\cite{liu2008isolation} recursively isolates anomalies through random splits, and (8) \textbf{SO\_GAAL}~\cite{liu2019generative} generates adversarial samples to identify anomalies. 
Reconstruction-based approaches include (9) \textbf{AE}~\cite{Aggarwal2017}, which flags anomalies based on reconstruction errors, and (10) \textbf{VAE}~\cite{kingma2013auto, burgess2018understanding}, which identifies anomalies using reconstruction probabilities or latent deviations. 
Finally, (11) \textbf{LUNAR}~\cite{goodge2022lunar} enhances traditional local outlier detection with graph neural networks.
\vspace{-0.1in}


\section{Experiment Results}
\vspace{-0.1in}
\label{sec:experiments}
\subsection{Experiment Setting}

\textbf{Datasets, Train/Test Data Split, and Independent Trials.} In the \method benchmark, the data is divided by allocating 70\% of the normal data to the training set. The remaining 30\% of normal data, combined with all anomalous data, forms the test set. To ensure the robustness of our findings, we repeat each experiment three times and report the average performance.

\noindent
\textbf{Hyperparameter Settings.} For all the algorithms in \method, we use their default hyperparameter (HP) settings in the original paper for a fair comparison, same as ADBench \cite{han2022adbench}.

\noindent
\textbf{Evaluation Metrics and Statistical Tests.} We evaluate different NLP-AD methods by a widely used metric: AUROC (Area Under Receiver Operating Characteristic Curve) and AUPRC (Area Under Precision-Recall Curve) value. 

\noindent
\textbf{Embeddings Definitions:} 
\begin{enumerate}[leftmargin=*, itemsep=0pt, parsep=0pt]
  \vspace{-0.1in}
  \item \textbf{BERT} refers specifically to the \textit{bert-base-uncased model} \cite{devlin2019bert}.
  \item \textbf{OpenAI} refers to OpenAI’s \textit{text-embedding-3-large} model \cite{openai2024embedding}.
  \item The term “BERT + AD algorithm” or “OpenAI + AD algorithm” means that we first generate text embeddings using BERT or OpenAI's model, respectively, and then apply the AD algorithm.
\end{enumerate}

\vspace{-0.2in}

\subsection{Results, Discussions, and New Directions}

We analyze the AUROC results presented in Table~\ref{tab:auroc} and the average rank summary in Figure~\ref{fig:auroc}. For completeness, AUPRC scores and their corresponding average ranks are reported in Appendix~\ref{sec:aupr-appendix}.
\vspace{-0.1in}

\def\doubleunderline#1{\underline{\underline{#1}}}
\begin{table*}
\caption{Performance comparison of 19 Algorithms on 8 NLPAD datasets using AUROC, with \textbf{best} results highlighted in \colorbox{gray!20}{\textbf{bold and shaded}}.}
\vspace{-0.1in}
\label{tab:auroc}
\centering
\small
\setlength{\tabcolsep}{3pt}
\renewcommand{\arraystretch}{0.9}
\fontsize{8}{10}\selectfont
\begin{tabular*}{\textwidth}{@{\extracolsep{\fill}}l >{\centering\arraybackslash}p{1.5cm} >{\centering\arraybackslash}p{1.5cm} >{\centering\arraybackslash}p{1.5cm} >{\centering\arraybackslash}p{1.5cm} >{\centering\arraybackslash}p{1.5cm} >{\centering\arraybackslash}p{1.5cm} >{\centering\arraybackslash}p{1.5cm} >{\centering\arraybackslash}p{1.5cm}}
\toprule
\textbf{Methods} & \textbf{NLPAD-AGNews} & \textbf{NLPAD-BBCNews} & \textbf{NLPAD-EmailSpam} & \textbf{NLPAD-Emotion} & \textbf{NLPAD-MovieReview} & \textbf{NLPAD-N24News} & \textbf{NLPAD-SMSSpam} & \textbf{NLPAD-YelpReview} \\ 
\midrule
CVDD & 0.6046 & 0.7221 & 0.9340 & 0.4867 & 0.4895 & 0.7507 & 0.4782 & 0.5345 \\
DATE & 0.8120 & 0.9030 & \colorbox{gray!20}{\textbf{0.9697}} & 0.6291 & 0.5185 & 0.7493 & \colorbox{gray!20}{\textbf{0.9398}} & 0.6092 \\
FATE* & 0.7756 & 0.9310 & 0.9061 & 0.5035 & 0.5289 & 0.8073 & 0.6262 & 0.5945 \\
\midrule
BERT + LOF & 0.7432 & 0.9320 & 0.7482 & 0.5435 & 0.4959 & 0.6703 & 0.7190 & 0.6573 \\
BERT + DeepSVDD & 0.6671 & 0.5683 & 0.6937 & 0.5142 & 0.4287 & 0.4366 & 0.5859 & 0.5871 \\
BERT + ECOD & 0.6318 & 0.6912 & 0.7052 & 0.5889 & 0.4282 & 0.4969 & 0.5606 & 0.6326 \\
BERT + iForest & 0.6124 & 0.6847 & 0.6779 & 0.4944 & 0.4420 & 0.4724 & 0.5053 & 0.5971 \\
BERT + SO-GAAL & 0.4489 & 0.3099 & 0.4440 & 0.5031 & 0.4663 & 0.4135 & 0.3328 & 0.4712 \\
BERT + AE & 0.7200 & 0.8839 & 0.4739 & 0.5594 & 0.4650 & 0.5749 & 0.6918 & 0.6441 \\
BERT + VAE & 0.6773 & 0.7409 & 0.4737 & 0.5594 & 0.4398 & 0.4949 & 0.6082 & 0.6441 \\
BERT + LUNAR & 0.7694 & 0.9260 & 0.8417 & 0.5186 & 0.4687 & 0.6284 & 0.6953 & 0.6522 \\
\midrule
OpenAI + LOF & 0.8905 & 0.9558 & 0.9263 & 0.7304 & 0.6156 & 0.7806 & 0.7862 & 0.8733 \\
OpenAI + DeepSVDD & 0.4680 & 0.5766 & 0.4415 & 0.4816 & 0.6563 & 0.6150 & 0.3491 & 0.5373 \\
OpenAI + ECOD & 0.7638 & 0.7224 & 0.9263 & 0.6206 & \colorbox{gray!20}{\textbf{0.7366}} & 0.7342 & 0.4317 & 0.5984 \\
OpenAI + iForest & 0.5213 & 0.6064 & 0.6937 & 0.5889 & 0.5064 & 0.4944 & 0.3751 & 0.5871 \\
OpenAI + SO-GAAL & 0.5945 & 0.2359 & 0.4440 & 0.5031 & 0.6201 & 0.5043 & 0.5671 & 0.5082 \\
OpenAI + AE & 0.8326 & 0.9520 & 0.7651 & 0.7067 & 0.6088 & 0.7155 & 0.5511 & 0.8524 \\
OpenAI + VAE & 0.8144 & 0.7250 & 0.5273 & 0.7067 & 0.4515 & 0.7418 & 0.4259 & 0.6163 \\
OpenAI + LUNAR & \colorbox{gray!20}{\textbf{0.9226}} & \colorbox{gray!20}{\textbf{0.9732}} & 0.9343 & \colorbox{gray!20}{\textbf{0.9328}} & 0.6474 & \colorbox{gray!20}{\textbf{0.8320}} & 0.7189 & \colorbox{gray!20}{\textbf{0.9452}} \\
\bottomrule
\end{tabular*}
\vspace{-0.2in}
\end{table*}

\begin{figure}[!th]
\centering
\includegraphics[width=\linewidth]{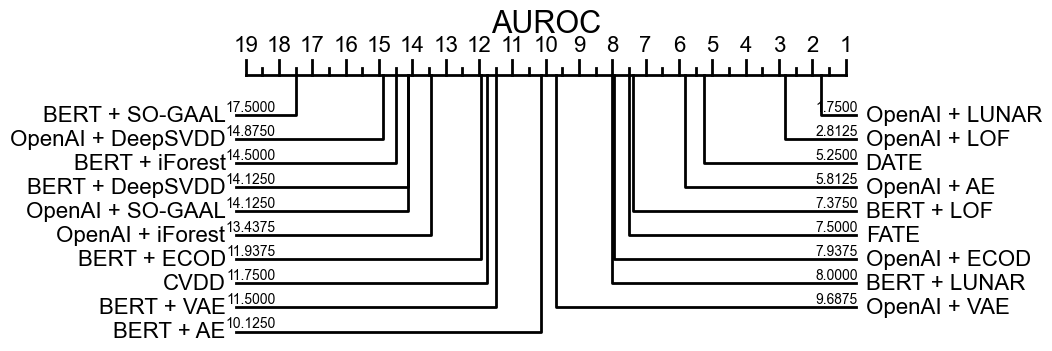} 
\vspace{-0.25in}
\caption{Average rank on AUROC of 19 NLPAD methods across 8 datasets (the lower the better).}
\vspace{-0.15in}
\label{fig:auroc}
\end{figure}

\noindent
\textbf{No single model consistently excels across all datasets due to variability in dataset characteristics.}  
AD model performance varies significantly across datasets, complicating the selection of a universally optimal model.
For datasets with more categories (e.g., NLPAD-AGNews), two-step methods like OpenAI + LUNAR (\textbf{0.9226}) outperform end-to-end methods such as CVDD (\textbf{0.6046}) by \textbf{52.6\%}. Similarly, on NLPAD-BBCNews, OpenAI + LOF (\textbf{0.9558}) surpasses CVDD (\textbf{0.7221}) by \textbf{32.4\%}. Conversely, on the binary-class datasets (e.g., NLPAD-SMSSpam), end-to-end methods perform better, with DATE (\textbf{0.9398}) clearly exceeding OpenAI + LUNAR (\textbf{0.7189}) by \textbf{30.7\%}.
\vspace{-0.1in}
\begin{itemize}[leftmargin=*]
    
    \item \textit{Future Direction 1: Automated Model Selection.} 
    These results emphasize the importance of developing automated approaches to select the most suitable model. One feasible solution will be adapting the meta-learning framework from tabular AD settings \cite{zhao2021automatic} to NLP-AD.
\end{itemize}
\vspace{-0.1in}

\noindent
\textbf{Transformer-based embeddings boost the performance of two-step AD methods.}  
Two-step AD algorithms paired with transformer-based embeddings have consistently outperformed end-to-end methods in NLP-AD tasks.
For instance, OpenAI + LUNAR achieves \textbf{0.9226} on NLPAD-AGNews, surpassing CVDD by \textbf{52.6\%} and FATE* by \textbf{19.0\%}. Similarly, OpenAI + LOF reaches \textbf{0.9558} on NLPAD-BBCNews, exceeding CVDD by \textbf{32.4\%} and FATE* by \textbf{2.7\%}. This advantage arises primarily because two-step methods leverage superior contextual embeddings from modern transformer models (e.g., OpenAI), whereas end-to-end methods like CVDD rely on older embeddings (e.g., GloVe). This highlights the need for end-to-end methods to adopt more advanced embeddings to enhance performance.

\begin{itemize}[leftmargin=*]
\vspace{-0.1in}
\item \textit{Future Direction 2: Transformer Embedding Integration for End-to-End AD.} Future end-to-end methods should adopt transformer-based embeddings over static embeddings like GloVe. Research should focus on embedding integration optimized for end-to-end AD frameworks.
\end{itemize}

\vspace{-0.1in}
\noindent 
\textbf{High-dimensional embeddings enhance detection but require balancing performance and efficiency.}
Embedding dimensionality significantly impacts both performance and computational efficiency in AD tasks. Compared to BERT-base embeddings (768 dimensions), OpenAI’s \textit{text-embedding-3-large} embeddings (3072 dimensions, a \textbf{300\% increase}) consistently achieve superior results across multiple datasets in \method. Specifically, OpenAI + LUNAR achieves \textbf{0.9452} on NLPAD-YelpReview (outperforming BERT + LUNAR’s \textbf{0.6522} by \textbf{44.9\%}), \textbf{0.9226} on NLPAD-AGNews (exceeding BERT + LUNAR’s \textbf{0.7694} by \textbf{19.9\%}), and \textbf{0.8320} on NLPAD-N24News (surpassing BERT + LUNAR’s \textbf{0.6284} by \textbf{32.4\%}). These results clearly demonstrate the advantage of higher-dimensional embeddings for enhancing AD performance. However, higher dimensionality also introduces greater computational costs and potential information redundancy.
\vspace{-0.1in}
\begin{itemize}[leftmargin=*]
    \item \textit{Future Direction 3: Optimizing Embedding Dimensionality.}
Future research should explore NLP-AD-specific dimensionality reduction techniques to reduce redundancy and computational costs without compromising performance. Additionally, adaptive methods that dynamically adjust dimensionality based on dataset characteristics could enhance scalability and efficiency.
\end{itemize}
\vspace{-0.2in}

\subsection{In-depth Analysis of Key Findings}
\vspace{-0.1in}
\subsubsection{Explaining Method-Dataset Fit}
\vspace{-0.05in}
To understand why certain models outperform others on specific datasets, we conduct both quantitative and qualitative analyses to identify dataset-level factors influencing model performance.

\noindent
\textbf{Quantitative Corpus-Level Linguistic Analysis} 
We characterize each dataset using three linguistic indicators:  
(1) \textbf{Avg-Len}, the average number of BERT tokens per sample (text complexity);  
(2) \textbf{Lexical-Burstiness}, the proportion of top-20 anomaly-specific tokens from BERT-tokenized TF-IDF (higher values indicate stronger lexical anomaly signals);  
(3) \textbf{Topic-Diversity}, the Shannon entropy over the label distribution of normal-class examples (0 for binary datasets; higher values indicate broader topic coverage).
\vspace{-0.1in}
\begin{table}[h]
\centering
\caption{Dataset characteristics and best method}
\vspace{-0.1in}
\label{tab:dataset_characteristics}
\resizebox{\columnwidth}{!}{%
\begin{tabular}{lccccc}
\toprule
Dataset & Avg-Len & Lexical-Burst. & Topic-Div. & Best Method \\
\midrule
NLPAD-AGNews & 39.5 & 0.040 & 1.585 & OpenAI+LUNAR (two-step)  \\
NLPAD-BBCNews & 481.7 & 0.028 & 1.981 & OpenAI+LUNAR (two-step)  \\
NLPAD-EmailSpam & 238.4 & 0.024 & 0.000 & DATE (end-to-end)  \\
NLPAD-Emotion & 20.3 & 0.087 & 1.949 & OpenAI+LUNAR (two-step)  \\
NLPAD-MovieReview & 290.3 & 0.026 & 0.000 & OpenAI+ECOD (two-step)  \\
NLPAD-N24News & 1034.2 & 0.016 & 4.437 & OpenAI+LUNAR (two-step)  \\
NLPAD-SMSSpam & 20.5 & 0.063 & 0.000 & DATE (end-to-end) \\
NLPAD-YelpReview & 151.6 & 0.023 & 0.000 & OpenAI+LUNAR (two-step)  \\
\bottomrule
\end{tabular}%
}
\end{table}
\vspace{-0.1in}

\noindent
From Table.~\ref{tab:dataset_characteristics}, we get two findings: (1) Datasets with short texts, high lexical-burstiness, and explicit lexical markers (e.g., NLPAD-SMSSpam) tend to favor DATE’s token-level anomaly detection. (2) Datasets with moderate length, lower lexical-burstiness, and high topical diversity (e.g., NLPAD-AGNews) benefit from richer, context-sensitive embeddings (OpenAI+LUNAR).

\noindent
\textbf{Qualitative Analysis of Representative Cases}
To further investigate the model performance gap, we qualitatively examined anomaly examples from \textit{NLPAD-SMSSpam} and \textit{NLPAD-AGNews} (see Table~\ref{tab:anomaly_examples}).  
In \textit{NLPAD-SMSSpam}, anomalies often contain explicit lexical irregularities, such as numeric tokens, unconventional formatting, or urgency-inducing phrases. These surface-level features align closely with DATE’s self-supervised scoring mechanism, which is sensitive to token-level deviations. By contrast, anomalies in \textit{NLPAD-AGNews} exhibit subtle semantic shifts without distinctive lexical markers. Detecting such anomalies requires a deeper understanding of contextual semantics, which exceeds DATE’s capacity and favors two-step methods with embedding-based models such as OpenAI and LUNAR.
\vspace{-0.1in}
\begin{table}[h]
\centering
\caption{Representative anomaly examples from NLPAD-SMSSpam and NLPAD-AGNews}
\vspace{-0.1in}
\label{tab:anomaly_examples}
\resizebox{\columnwidth}{!}{%
\begin{tabular}{p{9.5cm}p{5.5cm}}
\toprule
\textbf{NLPAD-SMSSpam} & \textbf{NLPAD-AGNews} \\
\midrule
PRIVATE! Your 2003 Account Statement for shows 800 un redeemed S. I. M. points. Call 08715203694 Identifier Code: 40533 Expires 31/10/04 & 
\multirow{2}{5.5cm}{NEW YORK U.S. stocks are expected to open modestly higher Tuesday as investors use a slight let up in the rise in oil prices to add to portfolios...} \\[3mm]

FREE for 1st week! No1 Nokia tone 4 ur mob every week just txt NOKIA to 87077 Get txting and tell ur mates. zed POBox 36504 W45WQ norm150p/tone 16 & 
 \\[3mm]

Urgent! Please call 09061213237 from landline. 5000 cash or a luxury 4 Canary Islands Holiday await collection. T Cs SAE PO Box 177. M227XY. 150ppm. 16 & \multirow{2}{5.5cm}{AFP Indian shares, Asia's second top performers last year, are poised for long term gains as foreign investors buy into the market, seeing the country as an economic "growth story," according to analysts.}\\[3mm]

XXXMobileMovieClub: To use your credit, click the WAP link in the next txt message or click here xxxmobilemovieclub.com?n QJKGIGHJJGCBL & \\
\bottomrule
\end{tabular}%
}
\vspace{-0.2in}
\end{table}

\subsubsection{Performance-Efficiency Trade-offs in Embedding Dimensionality}

While OpenAI-based two-step methods achieve high anomaly detection performance, their high dimensionality raises concerns about computational and financial cost in deployment scenarios. To explore whether such overhead is justified, we conduct dimensionality reduction experiments using PCA, projecting OpenAI embeddings to 768 dimensions—the same as BERT. 

\vspace{-0.1in}
\begin{table}[h]
\centering
\caption{Performance-efficiency trade-offs of dimensionality reduction on datasets}
\vspace{-0.1in}
\label{tab:dimensionality_reduction}
\resizebox{\columnwidth}{!}{
\begin{tabular}{lcccccccc}
\toprule
\multirow{2}{*}{Method} & \multirow{2}{*}{\makecell{Embedding Dim}} & \multicolumn{2}{c}{Performance} & \multicolumn{4}{c}{Runtime (s/sample)} \\
\cmidrule(lr){3-4} \cmidrule(lr){5-8}
& & AUROC & AUPRC & Total & Embedding & PCA & Inference \\
\midrule
\midrule
\multicolumn{8}{c}{\textbf{NLPAD-AGNews}} \\
OpenAI + LUNAR (orig.) & 3072 & \textbf{0.907} & \textbf{0.562} & 0.377 & 0.267 & 0.000 & 0.110 \\
OpenAI + LUNAR (PCA) & 768 & 0.890 & 0.546 & 0.409 & 0.252 & 0.007 & 0.150 \\
BERT + LUNAR & 768 & 0.790 & 0.325 & \textbf{0.081} & 0.056 & 0.000 & 0.025 \\
\midrule
OpenAI + LOF (orig.) & 3072 & \textbf{0.896} & \textbf{0.575} & 0.314 & 0.239 & 0.000 & 0.075 \\
OpenAI + LOF (PCA) & 768 & 0.798 & 0.321 & 0.435 & 0.245 & 0.007 & 0.183 \\
BERT + LOF & 768 & 0.771 & 0.304 & \textbf{0.076} & 0.062 & 0.000 & 0.014 \\
\midrule
\midrule
\multicolumn{8}{c}{\textbf{NLPAD-MovieReview}} \\
OpenAI + LUNAR (orig.) & 3072 & 0.664 & 0.238 & 0.322 & 0.270 & 0.000 & 0.052 \\
OpenAI + LUNAR (PCA) & 768 & \textbf{0.681} & \textbf{0.249} & 0.409 & 0.270 & 0.007 & 0.132 \\
BERT + LUNAR & 768 & 0.467 & 0.152 & \textbf{0.071} & 0.060 & 0.000 & 0.010 \\
\midrule
OpenAI + LOF (orig.) & 3072 & \textbf{0.652} & \textbf{0.242} & 0.314 & 0.274 & 0.000 & 0.041 \\
OpenAI + LOF (PCA) & 768 & 0.624 & 0.226 & 0.384 & 0.256 & 0.007 & 0.121 \\
BERT + LOF & 768 & 0.498 & 0.166 & \textbf{0.068} & 0.060 & 0.000 & 0.009 \\
\bottomrule
\end{tabular}
}
\end{table}
\vspace{-0.1in}

As shown in Table~\ref{tab:dimensionality_reduction}, PCA-reduced OpenAI embeddings slightly change performance (e.g., AUROC drops from 0.907 to 0.890 on NLPAD-AGNews with LUNAR) and consistently outperform BERT at the same dimension. However, PCA increases total runtime because it compresses embeddings into a denser space, which complicates decision boundaries and slows down inference.

This finding reinforces \textit{Future Direction 3}, highlighting the need for NLP-AD-specific dimensionality reduction techniques that balance representation quality with computational efficiency.
\vspace{-0.1in}
\section{Conclusion}
\vspace{-0.1in}
We present \method, the most comprehensive benchmark for contextual NLP anomaly detection (NLP-AD), evaluating 19 state-of-the-art algorithms across 8 diverse datasets. 
Our findings establish the superiority of two-step methods leveraging transformer-based embeddings, such as OpenAI + LUNAR, over end-to-end approaches, demonstrating the power of hybrid strategies for handling complex NLP anomaly detection tasks. By combining advanced text embeddings with traditional anomaly detection methods, \method provides a robust and flexible framework that sets a new standard for evaluating NLP-AD systems. 
Additionally, we offer actionable insights into model performance, dataset variability, and embedding utilization, paving the way for future research.
\vspace{1em}

\clearpage
\newpage
\section*{Limitations}
\vspace{-0.1in}
Despite its contributions, \method has certain limitations. First, the datasets included in the benchmark, while diverse, are primarily sourced from existing classification tasks and may not fully reflect emerging challenges such as anomalies in multilingual or multimodal text data. Second, our evaluations focus on static embeddings, leaving dynamic or streaming NLP-AD scenarios unexplored. 
Third, the reliance on predefined anomaly labels in our benchmark limits the ability to assess unsupervised or domain-adaptive approaches.
Future work can expand \method to include more diverse datasets, such as multilingual or multimodal data, and by exploring dynamic anomaly detection in streaming text scenarios. Incorporating benchmarks for unsupervised and adaptive models can also better reflect real-world applications. These advancements will enhance \method's utility as a comprehensive platform for driving progress in NLP anomaly detection.

\section*{Ethics Statement}
\vspace{-0.1in}
This work adheres to ethical standards emphasizing transparency, fairness, and privacy in NLP anomaly detection research. By openly sharing datasets, algorithms, and experimental results, \method provides a standardized foundation for advancing safer and more reliable web-based systems. All datasets are publicly available and contain no personally identifiable information, ensuring privacy compliance. Pre-trained embeddings (such as OpenAI’s text-embedding-3-large) are used in accordance with their terms of service. Additionally, we used ChatGPT exclusively to improve minor grammar in the final manuscript text.

\section*{Broader Impacts}
\vspace{-0.1in}

The NLP-ADBench proposed in this paper provides a comprehensive benchmark framework for anomaly detection in NLP. By standardizing datasets and algorithms, this work supports advancements in critical web-based applications, including fraud detection, spam filtering, and content moderation. The benchmark promotes transparency, reproducibility, and facilitates further innovations, ultimately contributing to safer, more reliable online environments.

\section*{Acknowledgments}
\vspace{-0.1in}
This work was partially supported by the National Science Foundation under Award Nos.~2428039, 2346158, and 2449280. 
We also acknowledge the use of computational resources provided by the Advanced Cyberinfrastructure Coordination Ecosystem~\cite{2c7a00c25fa845ad8a22b9298b756573}: Services \& Support (ACCESS) program, supported by NSF grants \#2138259, \#2138286, \#2138307, \#2137603, and \#2138296. Specifically, this work used NCSA Delta GPU at the National Center for Supercomputing Applications (NCSA) through allocation CIS250073. Any opinions, findings, conclusions, or recommendations expressed in this material are those of the authors and do not necessarily reflect the views of the National Science Foundation.
The authors also gratefully acknowledge support from the Amazon Research Awards and Capital One Research Awards.
\bibliography{sample-base,custom}

\clearpage
\newpage

\appendix

\section*{Supplementary Material for \method}

\setcounter{section}{0}
\setcounter{figure}{0}
\setcounter{table}{0}
\makeatletter 
\renewcommand{\thesection}{\Alph{section}}
\renewcommand{\theHsection}{\Alph{section}}
\renewcommand{\thefigure}{A\arabic{figure}} 
\renewcommand{\theHfigure}{A\arabic{figure}} 
\renewcommand{\thetable}{A\arabic{table}}
\renewcommand{\theHtable}{A\arabic{table}}
\makeatother

\renewcommand{\thetable}{A\arabic{table}}
\setcounter{equation}{0}
\renewcommand{\theequation}{A\arabic{equation}}

\section{Details on \method}

\subsection{Additional Details on Benchmark Datasets}
\label{appx:datasets}


\noindent
\subsubsection{Datasets Sources.}
\label{appx:datasets_sources}
\begin{enumerate}[leftmargin=*, itemsep=0pt, parsep=0pt]
    \item \textbf{NLPAD-AGNews} is constructed from the AG News dataset~\cite{rai2023agnews}, which was originally intended for news topic classification tasks. The AG News dataset contains 127,600 samples categorized into four classes: World, Sports, Business, and Sci/Tech. We selected the text from the “description” column as NLPAD-AGNews’s text data source. The “World” category was designated as the anomaly class and was downsampled accordingly. 

    \item \textbf{NLPAD-BBCNews} is constructed from the BBC News dataset \cite{greene06icml}, which was originally used for document classification across various news topics. The BBC News dataset includes 2,225 articles divided into five categories: Business, Entertainment, Politics, Sport, and Tech. We selected the full text of the news articles as NLPAD-BBC News’s text data source. The “Entertainment” category was designated as the anomaly class and was downsampled accordingly. 
    
    \item \textbf{NLPAD-EmailSpam} is constructed from the Spam Emails dataset~\cite{metsis2006spam}, originally used for email spam detection. The Spam Emails dataset contains 5,171 emails labeled as either spam or ham (not spam). We selected the text from the “text” column containing the email bodies as NLPAD-Emails Spam’s text data source. The “spam” category was designated as the anomaly class and was downsampled accordingly. 

    \item \textbf{NLPAD-Emotion:} is constructed from the Emotion dataset~\cite{saravia-etal-2018-carer} , which was originally intended for emotion classification tasks in textual data. The Emotion dataset contains 416,809 text samples labeled with six emotions: anger, fear, joy, love, sadness, and surprise. We selected the text from the “text” column as NLPAD-Emotion’s text data source. The “fear” category was designated as the anomaly class and was downsampled accordingly. 
    
    \item \textbf{NLPAD-MovieReview:} is constructed from the Movie Review dataset~\cite{maas-EtAl:2011:ACL-HLT2011} , commonly used for sentiment analysis of film critiques. The Movie Review dataset includes 50,000 reviews labeled as positive or negative. We selected the full review texts as NLPAD-MovieReview’s text data source. The “neg” (negative reviews) category was designated as the anomaly class and was downsampled accordingly. 
    
    \item \textbf{NLPAD-N24News} is constructed from the N24News dataset~\cite{wang-etal-2022-n24news}, originally used for topic classification of news articles. N24News contains 61,235 articles across various categories.
    We selected the full text of the news articles as NLPAD-N24News’s text data source. The “food” category was designated as the anomaly class and was downsampled accordingly.
    
    \item \textbf{NLPAD-SMSSpam} is constructed from the SMS Spam Collection dataset~\cite{10.1145/2034691.2034742}, originally intended for classifying SMS messages as spam or ham (not spam). The SMS Spam Collection dataset comprises 5,574 messages labeled accordingly. We selected the text from the “message text” as NLPAD-SMS Spam’s text data source. The “spam” category was designated as the anomaly class and was downsampled accordingly. 
    
    \item \textbf{NLPAD-YelpReview} is constructed from the Yelp Review Polarity dataset ~\cite{putra2023yelp}, originally intended for sentiment classification tasks. The Yelp Review Polarity dataset is created by considering 1-star and 2-star ratings as negative, and 3-star and 4-star ratings as positive. For each polarity, 280,000 training samples and 19,000 testing samples were randomly selected, resulting in a total of 560,000 training samples and 38,000 testing samples. Negative polarity is labeled as class 1, and positive polarity as class 2. We selected the text from the text column as NLPAD-YelpReview’s text data source. The label 1 (negative reviews) was designated as the anomaly class and was downsampled accordingly. 

\end{enumerate}

\subsubsection{NLPAD dataset's text pre-processing}
On all 8 datasets, we preprocessed the raw text data to ensure consistency and usability by removing URLs and HTML tags, eliminating unnecessary special characters while retaining essential punctuation, converting line breaks and consecutive spaces into single spaces, and preserving case sensitivity and stop words to maintain linguistic integrity. After processing the text, we found that some texts became duplicates due to the removal of certain symbols. Consequently, we removed all duplicate data to ensure the uniqueness of each text sample. These preprocessing steps follow established practices to effectively clean text data while retaining its syntactic and semantic features, providing a reliable foundation for natural language processing tasks~\cite{Chai2022Comparison}.
\subsection{Additional Details on Algorithms}
\label{appx:algorithms}
\subsubsection{End-to-End Algorithms} 
\begin{enumerate}[leftmargin=*, itemsep=0pt, parsep=0pt]
  \item \textbf{Context Vector Data Description}~\cite{ruff2019}(CVDD) is an unsupervised anomaly detection method for textual data. It utilizes pre-trained word embeddings and a multi-head self-attention mechanism to learn "context vectors" that represent normal patterns in the data. Anomalies are detected by measuring the cosine distance between sequence projections and context vectors, where larger distances indicate higher anomaly likelihoods. CVDD also penalizes overlapping contexts to enhance interpretability.

\item \textbf{Detecting Anomalies in Text via Self-Supervision of Transformers (DATE)}~\cite{manolache2021date} detects anomalies in text by training self-supervised transformers on tasks like replaced mask detection, enabling the model to learn normal language patterns and identify deviations.
\item \textbf{Few-shot Anomaly Detection in Text with Deviation Learning (FATE)}~\cite{das2023fate} is a deep learning framework that uses a small number of labeled anomalies to learn anomaly scores end-to-end. By employing deviation learning, it ensures normal examples align with reference scores while anomalies deviate significantly. Utilizing multi-head self-attention and multiple instance learning, FATE achieves state-of-the-art performance on benchmark datasets.  However, as our approach focuses on unsupervised anomaly detection, we adapt FATE into \textbf{FATE}* by training exclusively on normal data. This adaptation involves modifying the framework to learn reference scores and deviations without access to labeled anomalies, enabling effective detection of anomalous examples in an entirely unsupervised setting.

\end{enumerate}
\subsubsection{Traditional Algorithms}

\begin{enumerate}[leftmargin=*, itemsep=0pt, parsep=0pt]
  \item \textbf{Local Outlier Factor (LOF)}~\cite{breunig2000lof} calculates the local density deviation of a data point relative to its neighbors. This metric identifies points that have substantially lower density than their neighbors, marking them as outliers.
  \item \textbf{Deep Support Vector Data Description (DeepSVDD)}~\cite{ruff2018deep} minimizes the volume of a hypersphere enclosing the data representations learned by a neural network, capturing common patterns while identifying anomalies as points outside the hypersphere.
  \item \textbf{Empirical-Cumulative-distribution-based Outlier Detection (ECOD) }~\cite{li2022ecod} estimates the empirical cumulative distribution function (ECDF) for each feature independently. It identifies outliers as data points that reside in the tails of these distributions. This approach is hyperparameter-free and offers straightforward interpretability.
  \item \textbf{Isolation Forest (IForest)}~\cite{liu2008isolation} detects anomalies by isolating observations through random feature selection and splitting, with anomalies requiring fewer splits
  \item \textbf{Single-Objective Generative Adversarial Active Learning (SO\_GAAL)}~\cite{liu2019generative} optimizes a single objective function to generate adversarial samples and effectively identify anomalies in unsupervised settings.
  \item \textbf{AutoEncoder (AE)}~\cite{Aggarwal2017} detects anomalies by reconstructing input data, where higher reconstruction errors signify potential anomalies.
  \item \textbf{Unifying Local Outlier Detection Methods via Graph Neural Networks(LUNAR)}~\cite{goodge2022lunar} uses graph neural networks to integrate and enhance traditional local outlier detection methods, unifying them for better anomaly detection.
  \item \textbf{Variational AutoEncoder (VAE)}~\cite{kingma2013auto, burgess2018understanding} uses probabilistic latent variables to model data distributions, identifying anomalies based on reconstruction probabilities or latent space deviations.
\end{enumerate}

\subsection{More Experiment Results}
\begin{figure*}[!htbp]
\centering
\includegraphics[width=0.7\linewidth]{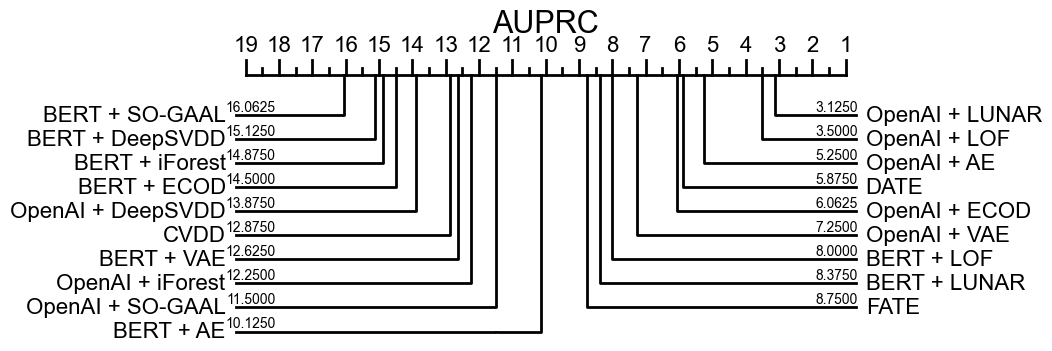}  
\caption{Average rank on AUPRC of 19 NLPAD methods across 8 datasets (the lower the better).}
\label{fig:auprc}
\end{figure*}

We also report AUPRC scores 
\label{sec:aupr-appendix}(Table.~\ref{tab:ad_results_auprc}) for all 19 algorithms across the 8 NLPAD datasets, along with their average AUPRC ranks (Fig.~\ref{fig:auprc}), to provide a complementary evaluation perspective beyond AUROC.

\begin{table*}[htbp]
\caption{Performance comparison of 19 Algorithms on 8 NLPAD datasets using AUPRC, with \textbf{best} results highlighted in \colorbox{gray!20}{\textbf{bold and shaded}}.}
\label{tab:ad_results_auprc}
\centering
\small
\setlength{\tabcolsep}{2pt}
\renewcommand{\arraystretch}{1}
\fontsize{8}{10}\selectfont
\begin{tabular*}
{\textwidth}{@{\extracolsep{\fill}} > {\raggedright\arraybackslash}p{2.6cm} >{\centering\arraybackslash}p{1.3cm} >{\centering\arraybackslash}p{1.3cm} >{\centering\arraybackslash}p{1.3cm} >{\centering\arraybackslash}p{1.3cm} >{\centering\arraybackslash}p{1.3cm} >{\centering\arraybackslash}p{1.3cm} >{\centering\arraybackslash}p{1.3cm} >{\centering\arraybackslash}p{1.3cm}}
\toprule
\textbf{Methods} & \textbf{NLPAD-AGNews} & \textbf{NLPAD-BBCNews} & \textbf{NLPAD-EmailSpam} & \textbf{NLPAD-Emotion} & \textbf{NLPAD-MovieReview} & \textbf{NLPAD-N24News} & \textbf{NLPAD-SMSSpam} & \textbf{NLPAD-YelpReview} \\ 
\midrule
CVDD & 0.1296 & 0.2976 & 0.5353 & 0.0955 & 0.1576 & 0.2886 & 0.0712 & 0.1711 \\
DATE & 0.3996 & 0.5764 & \colorbox{gray!20}{\textbf{0.8885}} & 0.1619 & 0.1682 & 0.2794 & \colorbox{gray!20}{\textbf{0.6112}} & 0.2149 \\
FATE* & 0.2787 & 0.5805 & 0.5529 & 0.1026 & 0.1752 & 0.2777 & 0.1257 & 0.2112 \\
\midrule
BERT + LOF & 0.2549 & 0.6029 & 0.2370 & 0.1170 & 0.1621 & 0.1678 & 0.1837 & 0.2629 \\
BERT + DeepSVDD & 0.2160 & 0.1328 & 0.2117 & 0.0986 & 0.1387 & 0.0798 & 0.1178 & 0.2174 \\
BERT + ECOD & 0.1616 & 0.2037 & 0.2077 & 0.1024 & 0.1374 & 0.0928 & 0.1156 & 0.2197 \\
BERT + iForest & 0.1559 & 0.2131 & 0.1894 & 0.1007 & 0.1412 & 0.0872 & 0.0994 & 0.2203 \\
BERT + SO-GAAL & 0.1033 & 0.0849 & 0.1130 & 0.1036 & 0.1486 & 0.0837 & 0.0714 & 0.2440 \\
BERT + AE & 0.2232 & 0.4274 & 0.2937 & 0.1037 & 0.1479 & 0.1255 & 0.1914 & 0.2525 \\
BERT + VAE & 0.1878 & 0.2559 & 0.2247 & 0.1019 & 0.1405 & 0.0957 & 0.1360 & 0.2331 \\
BERT + LUNAR & 0.2717 & 0.5943 & 0.3571 & 0.1053 & 0.1497 & 0.1436 & 0.1817 & 0.2609 \\
\midrule
OpenAI + LOF & 0.5443 & 0.7714 & 0.5967 & 0.2290 & 0.2133 & 0.2248 & 0.2450 & 0.5710 \\
OpenAI + DeepSVDD & 0.1062 & 0.1288 & 0.1195 & 0.1040 & 0.3278 & 0.1297 & 0.0721 & 0.1893 \\
OpenAI + ECOD & 0.3294 & 0.2424 & 0.5597 & 0.7443 & \colorbox{gray!20}{\textbf{0.5165}} & 0.2238 & 0.0821 & \colorbox{gray!20}{\textbf{0.8639}} \\
OpenAI + iForest & 0.1278 & 0.1376 & 0.3283 & 0.1311 & 0.1724 & 0.0913 & 0.0772 & 0.2527 \\
OpenAI + SO-GAAL & 0.1538 & 0.0665 & 0.1096 & 0.1291 & 0.3005 & 0.0963 & 0.1213 & 0.2735 \\
OpenAI + AE & 0.4022 & 0.7485 & 0.5580 & \colorbox{gray!20}{\textbf{0.8355}} & 0.1969 & 0.1984 & 0.1030 & 0.7063 \\
OpenAI + VAE & 0.3659 & 0.2424 & 0.5604 & 0.7744 & 0.1486 & 0.2537 & 0.0812 & 0.8467 \\
OpenAI + LUNAR & \colorbox{gray!20}{\textbf{0.6918}} & \colorbox{gray!20}{\textbf{0.8653}} & 0.5810 & 0.3112 & 0.2193 & \colorbox{gray!20}{\textbf{0.4425}} & 0.1640 & 0.4524 \\
\bottomrule
\end{tabular*}
\end{table*}

\end{document}